\documentclass[letterpaper, 10 pt, conference]{ieeeconf}  

\IEEEoverridecommandlockouts                              

\newif\ifcomment
\commenttrue   

\PassOptionsToPackage{usenames,dvipsnames}{color}
\usepackage{cite}
\usepackage{amsmath}
\usepackage{algorithm}
\usepackage{algpseudocode}
\usepackage{amsfonts}
\usepackage{pgfplots}
\pgfplotsset{compat=1.17}
\usepackage{hyperref}
\usepackage{booktabs}
\usepgfplotslibrary{groupplots} 
\usepackage[usenames, dvipsnames]{color}

\IEEEoverridecommandlockouts
\usepackage{tikz}
\usepackage{textcomp}
\usepackage{hyperref}
\usepackage{lipsum}

\newcommand\copyrighttext{%
    \footnotesize This work has been submitted to the IEEE for possible publication. Copyright may be transferred without notice, after which this version may no longer be accessible. 
 }
 
\newcommand\copyrightnotice{%
	\begin{tikzpicture}[remember picture,overlay]
		\node[anchor=south,yshift=10pt] at (current page.south) {\fbox{\parbox{\dimexpr\textwidth-\fboxsep-\fboxrule\relax}{\copyrighttext}}};
	\end{tikzpicture}%
}

\title{\LARGE \bf
Real-Time Whole-Body Control of Legged Robots \\ with Model-Predictive Path Integral Control
}

\author{Juan Alvarez-Padilla$^{1}$, John Z. Zhang$^{2}$, Sofia Kwok$^{2}$, John M. Dolan$^{2}$, and  Zachary Manchester$^{2}$
\thanks{$^{1}$Juan Alvarez-Padilla is with the Department of Electrical and Computer Engineering, Carnegie Mellon University.
        {\tt\small jralvare@andrew.cmu.edu}}%
\thanks{$^{2}$John Z. Zhang, Sofia Kwok, John Dolan, and Zachary  Manchester are with the Robotics Institute, Carnegie Mellon University.
        {\tt\small johnzhang@cmu.edu, sofiak@andrew.cmu.edu, jdolan@andrew.cmu.edu, zacm@cmu.edu}}%
}

\begin{document}
\maketitle
\ifcomment
    \copyrightnotice
\fi
\thispagestyle{empty}
\pagestyle{empty}
\begin{abstract}
This paper presents a system for enabling real-time synthesis of whole-body locomotion and manipulation policies for real-world legged robots. Motivated by recent advancements in robot simulation, we leverage the efficient parallelization capabilities of the MuJoCo simulator to achieve fast sampling over the robot state and action trajectories. Our results show surprisingly effective real-world locomotion and manipulation capabilities with a very simple control strategy. We demonstrate our approach on several hardware and simulation experiments: robust locomotion over flat and uneven terrains, climbing over a box whose height is comparable to the robot, and pushing a box to a goal position. To our knowledge, this is the first successful deployment of whole-body sampling-based MPC on real-world legged robot hardware. Experiment videos and code can be found at: 
\href{whole-body-mppi.github.io}{whole-body-mppi.github.io}
\end{abstract}

\section{Introduction}\label{sec: intro}
Building robots that can gracefully traverse difficult terrains and skillfully manipulate objects like humans and animals has been a long-standing goal in robotics. For a long time, the topics of robot locomotion~\cite{Ruina2005Efficient} and manipulation~\cite{osti_5761101} have been studied separately. Recent interest in general-purpose robot agents (i.e. humanoid robots) in both industry and academia has motivated designing control and planning algorithms capable of mastering both locomotion and manipulation skills in the same embodiment~\cite{he2024omnih2o, NVIDIA2024Foundation}. Despite this rise in interest, general methods capable of producing whole-body behaviors in real time on real-world quadruped and humanoid robots have so far remained elusive.

In this paper, we take advantage of the increasing performance of modern-day robotics simulation technology, in particular, the MuJoCo physics engine~\cite{todorov2012mujoco}, to compute real-time control policies on legged robots using model-predictive path integral control (MPPI)~\cite{williams2017model, williams2017mppi}. Contrary to the common belief that sampling approaches in high-dimensional tasks are computationally intractable, especially in real-time scenarios, we find that MPPI can be surprisingly effective at solving legged robot locomotion and manipulation tasks with a few simple design choices. To the best knowledge of the authors, this work represents the first time sample-based whole-body control has been successfully deployed on real-world legged robots. 
\begin{figure}[t]
    \centering
    \includegraphics[width=\linewidth]{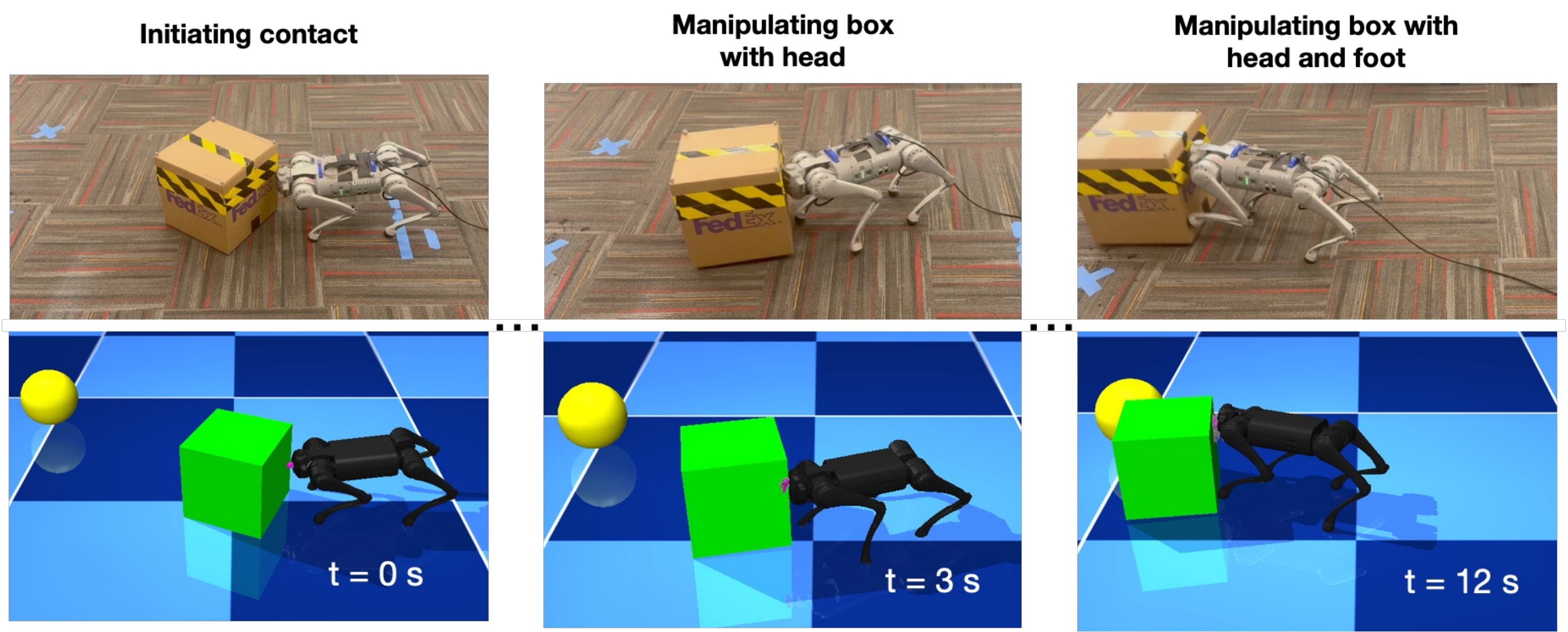}
    \caption{A Unitree Go1 robot pushing a box to a desired location with MPPI on hardware (top row) and corresponding MuJoCo simulation states (bottom row) on a single sequence. Contact-rich behaviors like body pushes and leg kicks emerge in real-time without manual pre-specification or offline policy training.}
    \label{fig:teaser}
\end{figure}

There are several key design choices in our implementation: First, we reduce the size of the search space by sampling over the control points of smooth splines in the robot's joint space and then tracking with low-level PD controllers to produce torque commands. Second, we leverage performant multi-threaded robot simulation to achieve fast real-time sampling and evaluation of simulation rollouts. Finally, we identify key controller parameters through both empirical observations and a set of ablation studies in controlled simulation environments. Our real-world and simulation results show that a sampling-based controller enabled by a modern physics engine can effectively reason about whole-body contact-rich behaviors during locomotion and manipulation that are challenging for gradient-based MPC algorithms. Additionally, different from RL approaches that require expensive offline training, our controller reasons about such contact-rich behaviors online in real time. 

Our specific contributions include:
\begin{enumerate}
    \item A system for deploying sampling-based predictive controllers on legged robots in real time.
    \item A set of ablation studies demonstrating the importance of algorithm hyperparameters that impact system performance.
    \item A collection of hardware and simulation experiments demonstrating our system's capabilities for solving high-dimensional, contact-rich, whole-body control problems in real time.
\end{enumerate}

\begin{figure*}[t]
    \centering
    \includegraphics[width=0.95\linewidth]{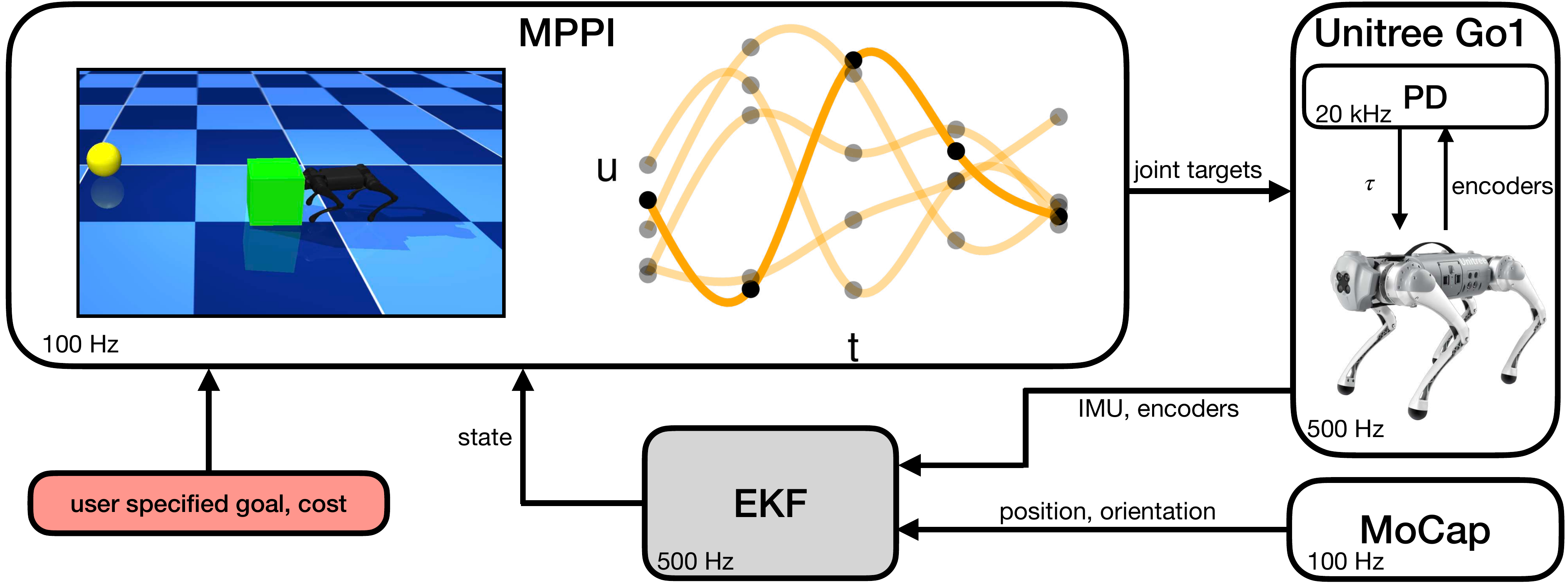}
    \caption{System diagram for deploying the MPPI policy on a Unitree Go1 robot. Joint target controls ($u$) are sampled at the evenly distributed knot points (black dots) and represented as a cubic spline over the planning horizon. A cost from each sample is evaluated based on the user-specified goal (yellow ball) and cost function. The first control from the control sequence with the lowest total cost (opaque orange line) is applied to the robot and repeated in a receding-horizon fashion. The robot's state is estimated using an EKF from motion-captured position and orientation, robot onboard IMU, and joint encoder measurements. }
    \label{fig:system_overview}
    \vspace{-5mm}
\end{figure*}
The remainder of this paper is organized as follows: We first review related works on legged robot control and MPPI in Section \ref{sec: background}. Next, we present our system design in Section \ref{sec: methods}. Then, we describe real-world results and ablation studies in simulation in Section \ref{sec: results}. Finally, we summarize our conclusions and point to avenues for future research in Section \ref{sec: conclusions}.

\section{Background and Related Work}\label{sec: background}
This section reviews related algorithms for legged robot locomotion and manipulation and relevant literature on MPPI.
\subsection{Locomotion and Manipulation for Legged Robots}
Current algorithms for legged robot locomotion and manipulation generally fall into two categories: gradient-based model-predictive control (MPC)~\cite{bishop2024reluqp, cleach2024fast, kuindersma2016optimization, bledt2018cheetah} and gradient-free reinforcement learning (RL)~\cite{lee2020learning, hwangbo2019learning,cheng2024parkour, pmlr-v205-margolis23a}. MPC policies leverage first or second-order gradient information from the model to achieve real-time policy optimization without any offline computation. Despite the obvious pitfalls of relying on simplified models, MPC has been a staple for real-world deployment of legged robots and demonstrates impressive generalization to different robots and surprising robustness to model mismatch, even on challenging terrains. However, due to real-time and onboard computation requirements for these algorithms, simplified models are often employed and only contacts between the feet and the terrain are considered~\cite{bledt2018cheetah, zhang2023slomo}, preventing these model-based policies from taking full advantage of the robot's dynamics and leveraging full-body contact to solve complicated locomotion or manipulation tasks. Existing approaches for model-based loco-manipulation require introducing manually designed task-specific contact pairs to the model~\cite{rigo2023contact,de_vincenti2023centralized}, significantly limiting generalization capabilities when facing new tasks.

On the other hand, simulation-based RL has shown impressive progress in recent years thanks to improved sim-to-real transfer via domain randomization techniques~\cite{lee2020learning, cheng2024parkour} and efficient parallel simulation on modern hardware~\cite{pmlr-v164-rudin22a, makoviychuk2021isaac, mujoco_mjx}. The fundamental difference between simulation-based RL and MPC is that RL attempts to learn a neural network policy \textit{offline} through large-scale trial-and-error in simulation. In this offline policy optimization regime without real-time computation constraints, it is standard to train policies by simulating whole-body robot dynamics and collision geometries from all parts of the robot and its environment, enabling discovery of non-trivial contact modes while solving complicated locomotion and manipulation tasks~\cite{hoeller2024anymal, ji2023driblebot}. In the online setting, the optimized policy network can be evaluated at real-time rates on onboard computers~\cite{pmlr-v205-margolis23a, cheng2024parkour}. Compared to online MPC, one downside of this offline policy optimization approach is the significant computation and time required to find performant reward functions and hyperparameters for each task.

In this paper, we aim to combine the benefits of both online MPC and simulation-based RL by solving whole-body motion-planning and control problems in real time by directly sampling over whole-body dynamics and collision models, \textit{without} any offline policy optimization.
\begin{figure*}[t]
    \centering
    \includegraphics[width=\linewidth]{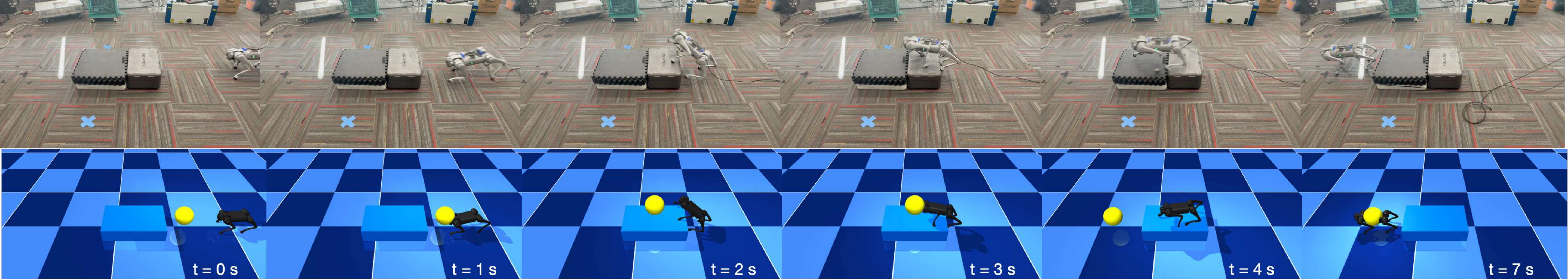}
    \caption{Keyframes from the Unitree Go1 robot climbing up and down a box of its own height with the MPPI policy on hardware (top row) and corresponding MuJoCo states (bottom row). The robot is tasked to reach the consecutive goals (yellow spheres) specified in the task.}
    \label{fig: box_climb}
    \vspace{-5mm}
\end{figure*}
\subsection{Model-Predictive Path Integral Control}
Model-predictive path integral (MPPI) control is a gradient-free sampling-based algorithm often applied to real-time motion planning and control. MPPI samples $N$ control trajectories from a multivariate Gaussian distribution $u_t \sim \mathcal{N}(\mu_t, \Sigma_t)$, where $\mu_t$ is the mean at time $t$ and $\Sigma_t$ is the covariance matrix. Each control trajectory is then simulated to compute a corresponding state trajectory, and a cost function is evaluated for each state-input trajectory sample. The control input to the system is then calculated through the exponentially weighted average of the samples based on their cost:
\begin{align}\label{eq:mppi_weights}
    \omega_n = \frac{\exp\left(-\frac{\mathcal{L}_n - \mathcal{L}_{\min}}{\lambda}\right)}{\sum_{n=1}^{N} \exp\left(-\frac{\mathcal{L}_n - \mathcal{L}_{\min}}{\lambda}\right)} ,
\end{align}
\begin{align} \label{eq:mppi_mu_update}
    \mu_t = \sum_{n=1}^{N} \omega_n u_{t},
\end{align}
where $\mathcal{L}_n$ represents the cost of the $n$-th trajectory, $\omega_n$ denotes the corresponding trajectory weight, and $\lambda$ is the temperature parameter that determines the controller's sensitivity to differences in trajectory cost. A lower value of $\lambda$ increases the influence of the best-performing trajectory, while a higher value distributes the weight more uniformly across all samples~\cite{williams2017model}. Finally, the process is repeated in a receding horizon approach. MPPI is summarized in Algorithm \ref{alg:mppi_algorithm}.

\begin{algorithm}[t]
\caption{MPPI Control Algorithm}
\begin{algorithmic}[1] 
\Require Initial state $x_0$, control parameters $(\mu_0, \Sigma_0)$, number of samples $N$, time horizon $T$, temperature parameter $\lambda$
\Ensure Optimal control input $u$
\State Initialize control trajectory mean $\mu_t$ and covariance $\Sigma_t$
\State Initialize sample trajectory cost $\mathcal{L}_n$ = 0
\For{each sample $n = 1, \dots, N$}
    \For{each timestep $t = 0, 1, \dots$}
        \State Sample action sequence: $ u_t \sim \mathcal{N}(\mu_t, \Sigma_t)$
        \State Simulate forward for $T$ timesteps using $ u_n $
        \State Compute the resulting cost and add it to $ \mathcal{L}_n $
    \EndFor
\EndFor
\State Compute weights \( \omega_n \) for each sample using Equation \ref{eq:mppi_weights}

\State Update control trajectory mean \( \mu_t \) using Equation \ref{eq:mppi_mu_update}

\State Select the control input for the system: $ u = \mu_t[0] $

\State Apply control input $u$
\State Shift control mean: $\mu_t \gets \text{shifted}(\mu_t)$

\end{algorithmic} \label{alg:mppi_algorithm}
\end{algorithm}

Driven by significant advancements in parallel computation on modern GPUs and its downstream effects on massively parallel simulation, MPPI has seen growing popularity and success in recent years~\cite{williams2017mppi, williams2016aggresive}. In off-road autonomous driving~\cite{williams2017mppi, williams2016aggresive} domains where environment dynamics are difficult to model analytically, MPPI naturally incorporates learned black-box models from real-world data thanks to its derivative-free nature, which is difficult for gradient-based MPC. However, so far, MPPI is mainly successfully deployed in low-dimensional control tasks (e.g., driving and drone flight) due to the curse of dimensionality for sampling-based algorithms.

Recent work~\cite{howell2022mjpc} proposes a simple yet effective modification better scale MPPI to higher-dimensional tasks (e.g., quadruped locomotion or dexterous manipulation): sample over the control points of a polynomial spline and calculate the remaining controls via spline interpolation. While this approach yielded impressive behaviors in simulation~\cite{howell2022mjpc}, closing the sim-to-real gap in real-world systems remains an open problem. Alternatively, \cite{carius2022constrained} deploys MPPI on a quadruped robot by learning a sampling distribution offline but only considers a kinodynamic robot model with foot contacts. Our work considers the whole-body dynamics and collision model, enabling automatic planning over contact strategies while solving locomotion and manipulation tasks without offline precomputation.

Rather than innovating over the MPPI algorithm itself, our work instead focuses on the system-level considerations and integration necessary to deploy this sampling-based control strategy on real-world legged robots that are high-dimensional, highly agile, and must reason about making and breaking contact with the environment.

\section{Sampling-based MPC for Legged Robots}\label{sec: methods}
This section discusses key implementation details and design considerations for deploying sampling-based MPC on real-world legged robot hardware. A diagram of the overall system is illustrated in Fig. \ref{fig:system_overview}.
\subsection{MuJoCo Physics Engine}
Robotics simulators are increasingly accessible and performant. In particular, Drake~\cite{drake} and Dojo~\cite{howelllecleach2022} focus on physical accuracy for algorithm verification and model-based control but pay the price of high computation costs and cannot generate simulation data at scale. Isaac~\cite{makoviychuk2021isaac} and MuJoCo~\cite{todorov2012mujoco,mujoco_mjx} provide mature parallel implementations and, as a result, have become popular choices for data-hungry algorithms like RL. For sampling-based MPC, the simulator of choice must generate large numbers of simulation samples in parallel while providing fast enough individual rollouts to close the real-time feedback control loop. With this consideration in mind, we use the MuJoCo physics engine and parallelize the rollouts on a multi-core CPU. While Isaac and MJX~\cite{mujoco_mjx} (MuJoCo on GPU or TPU) are capable of simulating hundreds to thousands of parallel rollouts compared to dozens for the CPU-based MuJoCo, the individual rollout speeds are insufficient to close the real-time feedback loop. In practice, we find that $30$-$50$ parallel rollouts in MuJoCo on a high-end CPU with strong per-core performance (e.g. recent Intel Core-i9) are sufficient to solve the locomotion and manipulation tasks considered in this paper. 

\subsection{Representing Controls as a Cubic Spline}
Directly sampling in the robot joint space can be challenging for MPPI, especially when planning over medium to long horizons. Following~\cite{howell2022mjpc}, we reduce the size of the search space by sampling over spline control points and interpolating using a cubic spline (Fig. \ref{fig:system_overview}). This spline representation also provides the added benefit of smoothing the controls. While it is possible to use zeroth-order or linear interpolation, we find that cubic splines perform the best in practice and focus on this representation in this paper. Similarly, in many applications, one can further reduce the search space by sampling over a reduced action space. in Sec. \ref{sec: ablations}, we investigate the impact of sampling representation on control performance.

\subsection{State Estimation}
We estimate the full state of the robot (global position, attitude, joint angles, body angular velocity, and joint velocities) by fusing position and attitude measurements from a motion-capture system ($100$ Hz), onboard IMU ($500$ Hz), and onboard joint encoders ($500$ Hz) with an EKF running at $500$ Hz, Fig. \ref{fig:system_overview}. Specifically, we take body velocity estimation from the single-rigid body EKF~\cite{bledt2018cheetah} and fuse it with the motion-capture measurements.
Different from standard locomotion policies~\cite{bledt2018cheetah, pmlr-v205-margolis23a} focusing on body-velocity estimation and tracking, precise global states are key to correctly resetting the MuJoCo simulation. Admittedly, a mismatched model between planning (whole-body, soft contact) and estimation (single-rigid body, hard contact) is not ideal and can degrade real-world performance. We find that this mismatch can be minimized with estimator tuning (especially the height of the robot, as it directly relates to planned contact forces during locomotion) and plan to further investigate estimation algorithms using MuJoCo dynamics in the future.


\section{Experiments and Results}\label{sec: results}
In this section, we present results demonstrating the capabilities of our sampling-based MPC implementation on a variety of tasks and describe the implementation details of the hardware system. In particular, we highlight the ability of our system to generate emergent whole-body contacts in real time while solving challenging locomotion and manipulation tasks without any contact pre-specification or offline policy optimization. Experiment videos and code can be found on our project website:
\begin{center}
\href{whole-body-mppi.github.io}{whole-body-mppi.github.io}
\end{center}

\subsection{Hardware Implementation}
We run MPPI on a workstation computer equipped with an Intel i9-12900KS CPU and 64 GB of memory. This workstation is connected to the robot via Ethernet and communicates using ROS. All hardware experiments are run on a Unitree Go1 quadruped robot. Across all hard experiments, we update the MPPI policy at $100$ Hz, discretize the dynamics at $0.01$ s, plan over a $0.4$ s horizon ($T = 40$ timesteps), and roll out $30$ samples. For contact simulation in MuJoCo, we use the default parameters for the contact model and rely on the linearized friction cone for speed. We refer the interested reader to our \href{whole-body-mppi.github.io}{website} for a full list of hyperparameters for each task. We publish joint targets from MPPI to the robot that are tracked using the Go1's onboard low-level PD controller at $20$ kHz.
\begin{figure}[t]
    \centering
    \includegraphics[width=\linewidth]{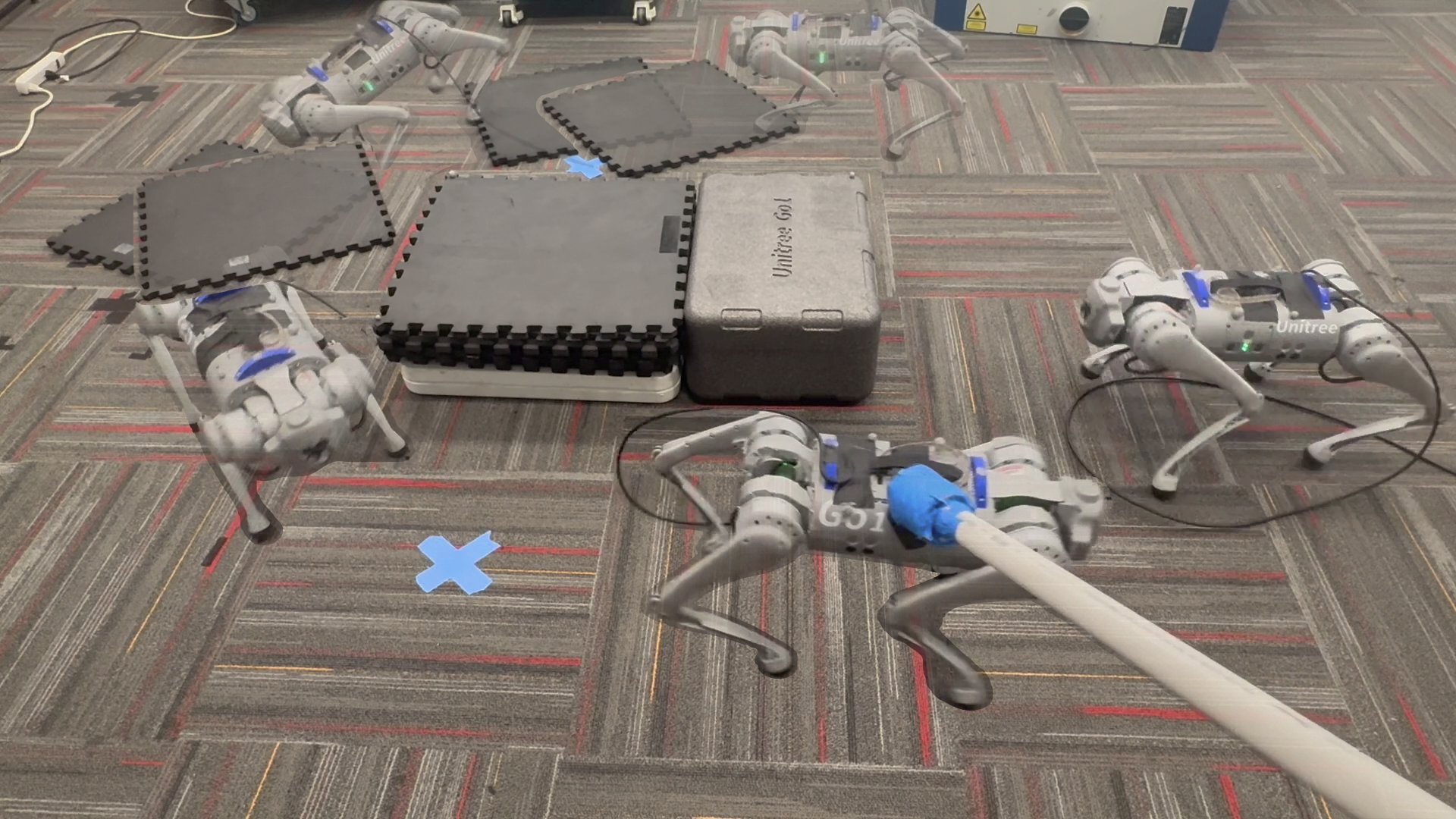}
    \caption{Go1 robot walking in a clockwise hexagon trajectory under small to moderate model mismatch and external disturbance. More transparent robots represent earlier keyframes.}
    \label{fig:walking}
    \vspace{-5mm}
\end{figure}

\subsection{Walking on Flat Terrain}
We first verify our implementation on a flat terrain locomotion task. The walking task cost function tracks a state and control reference:
\begin{align}
    \mathcal{L}_{\text{walk}} = \sum_{t=0}^{T} \Big[ (x_{\text{ref}} - x_t)^T Q (x_{\text{ref}} - x_t) + \nonumber \\
    (u_{\text{ref}}(t) - u_t)^T R (u_{\text{ref}}(t) - u_t) \Big] ,
\end{align}
where $x_t$ and $u_t$ are the state and control of the robot at index $t$, respectively. $x_{\text{ref}}$ includes the desired robot position and attitude. $u_{\text{ref}}(t)$ represents the joint targets of a walking gait based on the Raibert heuristic~\cite{raibert_dynamically_nodate}. Although we observe various locomotion gaits emerge by setting a time-invariant $u_{\text{ref}}$ to the standing pose and a goal position for the robot, we find the walking policy is significantly more robust when tracking a gait reference. Finally, $Q$ and $R$ are diagonal weight matrices on states and controls.

We successfully deploy the MPPI walking policy by trotting in place and walking to various user-specified waypoints (Fig. \ref{fig:walking}). Our policy is robust to moderate external disturbances and unmodeled terrain mismatch.

\subsection{Locomotion Over Challenging Terrains}
In more challenging locomotion scenarios, we model the terrain geometry in MuJoCo and set waypoints for the robot to track. We successfully deploy the MPPI policy on hardware to climb up a box of up to $0.24$ m tall --- roughly the height of the robot when standing (Fig. \ref{fig: box_climb}). Although we use the same trotting gait reference as when walking on flat ground, we observe the robot leverage unplanned motions such as jumps and contacts with the body to traverse the tall obstacle. 

\begin{figure}[t]
    \centering
    \includegraphics[width=0.9\linewidth]{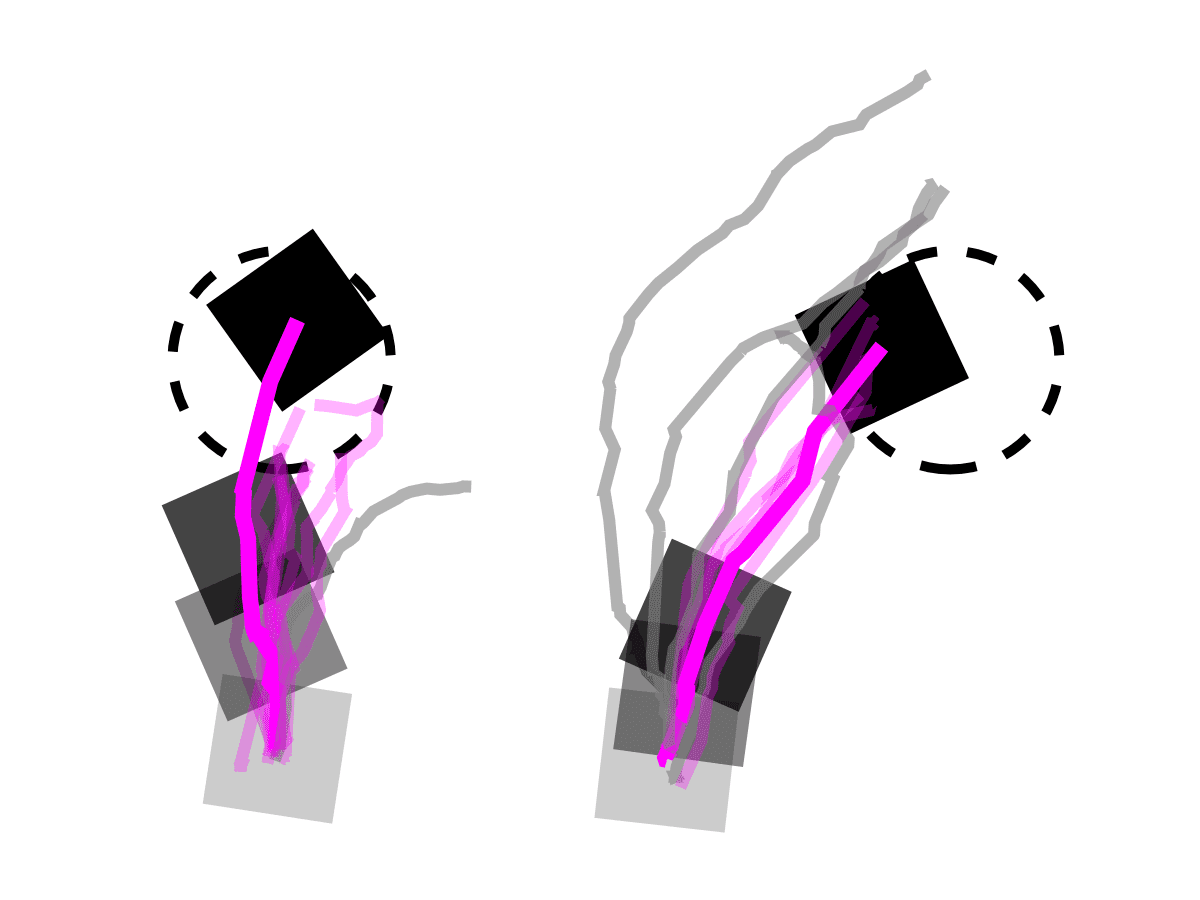}
    \vspace{-7mm}
    \caption{Top-down view of real-world box trajectories (magenta and grey lines) from $10$ trials of the Go1 robot pushing the box (black square) into the goal area (dashed circle) that is placed in front (left) and front-right (right) of the original box position. More transparent boxes represent earlier in the trajectory. Magenta lines represent runs where the box successfully reaches the target area while the grey lines indicate otherwise.}
    \label{fig: box_push}
    \vspace{-5mm}
\end{figure}

\subsection{Box Pushing}
Next, we showcase the whole-body contact planning capabilities of our MPPI policy to push a $3.5$ kg box of size $0.36\text{m}\times0.36\text{m}\times0.36\text{m}$, Fig. \ref{fig:teaser}. In addition to the locomotion cost on the robot, we add an $\ell_1$-norm cost for the position of the box: $\mathcal{L}_{\text{box}} = Q_{\text{box}} \lVert \Bar{x}_\text{box} - x_{\text{box}} \rVert_1$, where $\Bar{x}_\text{box}$ and $x_{\text{box}}$ are the target and current positions of the box. The total cost function for this task is:
\begin{align}
     \mathcal{L}_{\text{box push}} =  \mathcal{L}_{\text{walk}} +  \mathcal{L}_{\text{box}}
\end{align}
Note that while the box orientation feedback is important for planning, we do not penalize the orientation for this task. As a simple heuristic to encourage robot-box interaction, we place the goal for the robot at the center of the box but do not specify when or how the robot should interact with the object in any way.

We set the box $1$m in front of the quadruped and ask the robot to manipulate the box to two different goal locations: 1) directly $1$m forward and 2) $1$m forward and $0.75$m to the right from the original box position, Fig. \ref{fig:teaser}. We define a trial as a success and stop the run if the position of the box is within $0.3$m from the goal at any time. In the first scenario, we complete the task $9$ out of $10$ times. In the second, more challenging, scenario, we complete the task $6$ out of $10$ times. In this scenario, the robot must horizontally manipulate the box, requiring more sophisticated interactions between the robot body, leading to a larger sim-to-real gap. Box trajectories can be found in Fig. \ref{fig: box_push}. Experiment videos can be found on our \href{whole-body-mppi.github.io}{website} and in the supplementary video.

Interestingly, from online sampling in the robot action space, we observe emergent contact interactions between the robot and the box, such as body or shoulder pushes and leg kicks to move the box to the goal position.

\begin{figure}[t]
    \centering
    \begin{tikzpicture}
\definecolor{mediumpurple152142213}{RGB}{152,142,213}
\definecolor{steelblue52138189}{RGB}{52,138,189}

  \begin{axis}[
    ybar,
    bar width=16pt,
    enlarge y limits=false,
    legend style={
      at={(1,1)},
      anchor=north east,
      legend columns=1,
      draw=none,
      /tikz/align=left, 
    },
    legend image code/.code={
      \draw[#1] (0cm,-0.1cm) rectangle (0.3cm,0.1cm);
    },
    ylabel={Cost},
    ylabel style={
        at={(axis description cs:-0.1,1.0)}, 
        anchor=south, 
    },
    y label style={
        font=\small,
    },
    symbolic x coords={Direct, Zeroth-Order, Linear, Cubic},
    xtick=data,
    xtick style={draw=none},
    x tick label style={font=\small, yshift=-5pt},
    ymin=0,
    ymax=52e6,
    axis lines=box,
    clip=false,
    scaled y ticks=false, 
    ytick={0,10e6,20e6,30e6,40e6,50e6},
    yticklabels={0,10,20,30,40,50},
    y tick label style={
        /pgf/number format/.cd,
        fixed,
        precision=0,
    },
    extra description/.code={
      \node at (axis description cs:0.05,1.02) [anchor=south] {\small$\times10^{6}$};
    },
  ]
    \addplot+[
      ybar,
      fill=steelblue52138189,
      draw=black,
      line width=0.5pt,
      bar shift=-8pt,
      error bars/.cd,
        y dir=both,
        y explicit,
        error bar style={black},
        error mark=|,
        error mark options={
          draw=black
        },
    ] coordinates {
      (Direct,6.354e6) +- (0,1.8099e6)
      (Zeroth-Order,1.425e6) +- (0,77141.97)
      (Linear,1.321e6) +- (0,49614.70)
      (Cubic,1.034e6) +- (0,59539.94)
    };
    \addlegendentry{Trotting}
    \addplot+[
      ybar,
      fill=mediumpurple152142213,
      draw=black,
      line width=0.5pt,
      bar shift=8pt,
      error bars/.cd,
        y dir=both,
        y explicit,
        error bar style={black},
        error mark=|,
        error mark options={
          draw=black
        },
    ] coordinates {
      (Direct,44.056e6) +- (0,6.2915e6)
      (Zeroth-Order,20.213e6) +- (0,2.2211e6)
      (Linear,16.280e6) +- (0,4.9233e6)
      (Cubic,11.753e6) +- (0,4.7512e6)
    };
    \addlegendentry{Box Climb}
  \end{axis}
\end{tikzpicture}
    \caption{Policy rollout costs (lower is better) with different sampling representation for MPPI on trotting forward (blue) and box-climb (purple) tasks in MuJoCo. Directly sampling controls over the prediction horizon (Direct) performs significantly worse than sampling spline control points and interpolating. Cubic interpolation (ours) outperforms zeroth-order and linear interpolation in both tasks.}
    \label{fig: ablation_spline}
    \vspace{-5mm}
\end{figure}
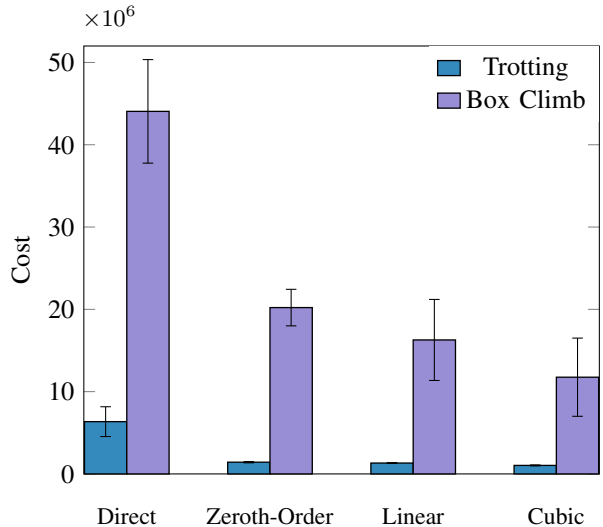

\begin{figure*} 
    \centering
    \begin{tikzpicture}
        \begin{groupplot}[
            group style={
                group size=4 by 1, 
                horizontal sep=0.12cm, 
                vertical sep=1cm, 
            },
            width=5.5cm, 
            height=5cm, 
            grid=major, 
            xlabel={X-axis}, 
            ymin=600000, ymax=1700000, 
            error bars/.cd]
        ]

        \nextgroupplot[ylabel={Cost}, xlabel={Control Frequency (Hz)}]
        
        \addplot+[
            mark=*, 
            color=blue,
            error bars/.cd,
            y dir=both, 
            y explicit 
        ] table [y error plus=UPPER, y error minus=LOWER] 
        {./plots/mujoco_traj_param_rate_10-250_h_40_lam_0.1_n_30_time_3s_task_walk_straight_avg_over_10_seeds_error.dat}; 

        \addplot+[
             mark=*,
            color=red,
            mark options={solid, color=red},
            error bars/.cd,
            y dir=both,
            y explicit
        ] table [y error plus=UPPER, y error minus=LOWER] 
        {./plots/gazebo_traj_param_rate_10-250_h_40_lam_0.1_n_30_time_3s_task_walk_straight_gz_avg_over_10_seeds_error.dat}; 
        \draw[dashed, black!80!blue, very thick] (axis cs:100,0) -- (axis cs:100,1700000);
        
        \legend{MuJoCo, Gazebo}
        
        \nextgroupplot[yticklabels={}, xlabel={Temperature $\lambda$}]

        \addplot+[
            mark=*, 
            color=blue,
            error bars/.cd,
            y dir=both, 
            y explicit 
        ] table [y error plus=UPPER, y error minus=LOWER] 
        {./plots/mujoco_traj_lambda_0.005-0.3_ctrl_ur_100_h_40_n_30_time_3s_task_walk_straight_avg_over_10_seeds_error.dat}; 

        \addplot+[
             mark=*,
            color=red,
            mark options={solid, color=red},
            error bars/.cd,
            y dir=both,
            y explicit
        ] table [y error plus=UPPER, y error minus=LOWER] 
        {./plots/gazebo_traj_lambda_0.005-0.3_ctrl_ur_100_h_40_n_30_time_3s_task_walk_straight_gz_avg_over_10_seeds_error.dat}; 
        \draw[dashed, black!80!blue, very thick] (axis cs:0.1,0) -- (axis cs:0.1,1700000);
        \nextgroupplot[yticklabels={}, xlabel={Prediction Horizon (timesteps)}]

        \addplot+[
            mark=*, 
            color=blue,
            error bars/.cd,
            y dir=both, 
            y explicit 
        ] table [y error plus=UPPER, y error minus=LOWER] 
        {./plots/mujoco_traj_horizon_range_5-100_ctrl_ur_100_lam_0.1_n_30_time_3s_task_walk_straight_avg_over_10_seeds_error.dat}; 

        \addplot+[
             mark=*,
            color=red,
            mark options={solid, color=red},
            error bars/.cd,
            y dir=both,
            y explicit
        ] table [y error plus=UPPER, y error minus=LOWER] 
        {./plots/gazebo_traj_horizon_range_5-100_ctrl_ur_100_lam_0.1_n_30_time_3s_task_walk_straight_gz_avg_over_10_seeds_error.dat}; 
        \draw[dashed, black!80!blue, very thick] (axis cs:40,0) -- (axis cs:40,1700000);

        \nextgroupplot[yticklabels={}, xlabel={Number of samples}]

        \addplot+[
            mark=*, 
            color=blue,
            error bars/.cd,
            y dir=both, 
            y explicit 
        ] table [y error plus=UPPER, y error minus=LOWER]{./plots/mujoco_traj_param_n_samples_range_5-500_h_40_lam_0.1_ctrl_100_time_3s_task_walk_straight_avg_over_10_seeds_error.dat}; 

        \addplot+[
             mark=*,
            color=red,
            mark options={solid, color=red},
            error bars/.cd,
            y dir=both,
            y explicit
        ] table [y error plus=UPPER, y error minus=LOWER]{./plots/gazebo_traj_n_samples_range_10-100_h_40_lam_0.1_ctrl_100_time_3s_task_walk_straight_gz_avg_over_10_seeds_error.dat}; 
        \draw[dashed, black!80!blue, very thick] (axis cs:30,0) -- (axis cs:30,1700000);
        \end{groupplot}
    \end{tikzpicture}
    \caption{Sim-to-sim policy rollout costs while varying key MPPI hyperparameters in both MuJoCo (blue) and Gazebo (red) simulation environments. The means and standard deviations are computed from $10$ different random seeds for each setting. Black dotted lines denote the default parameters we deploy on hardware.}
    \label{fig: ablations}
    \vspace{-5mm}
\end{figure*}
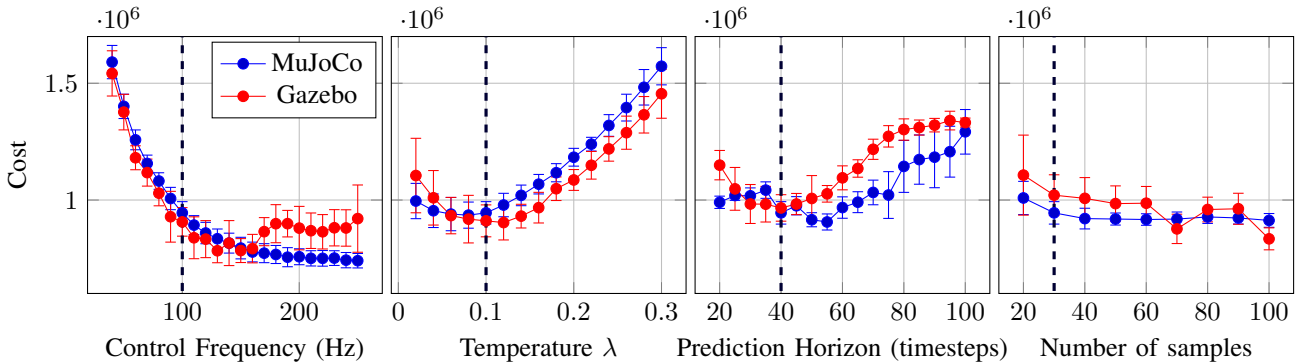

\subsection{Ablation Studies}\label{sec: ablations}
We investigate the effects of key MPPI hyperparameters on overall task performance in a controlled simulation setting. Note that it is nearly impossible to discuss a single hyperparameter's impact on overall system performance without the task or other parameters. Our goal is to provide an intuitive understanding of the tradeoffs when making design decisions. For all experiments, we fix the remaining hyperparameter to the default values presented above. In controlled MuJoCo and Gazebo simulation environments, we slow down simulation to match the policy update frequency to avoid real-time computation-related issues.
\subsubsection{Sampling representation}
We first look into the choice of sampling representation in MPPI on two tasks: walking forward and climbing up a box in MuJoCo (Fig. \ref{fig: ablation_spline}). Unsurprisingly, directly sampling robot joint targets over the entire prediction horizon is unable to sufficiently explore the optimization landscape given fixed samples compared to compressed sampling representations. Among interpolation schemes, higher-order polynomials generally produce better results, especially on a challenging task like climbing up a box. Given that we choose to sample four points along each trajectory, a cubic polynomial is the highest order possible. We leave higher-order interpolation in the presence of added knot points for future work.

The remaining ablations are performed on the walking task alone in both MuJoCo and sim-to-sim transfer to Gazebo:
\subsubsection{Control frequency}
Similar to other MPC algorithms, the controller update frequency is critical for deploying MPPI on legged robots. While the conventional wisdom of ``faster is better" still applies in this context, we find that performance plateaus after $\sim100$ Hz in Gazebo and $\sim150$ Hz in MuJoCo (Fig.~\ref{fig: ablations}). We believe this is due to our choice of sampling over joint targets and relying on low-level PD controllers at much higher frequencies to compute torques. In practice, we find $\sim100$ Hz to be sufficient on hardware.
\subsubsection{Temperature}
Temperature $\lambda$ is a key parameter in the MPPI algorithm. Lower $\lambda$ means added weight on the best-performing predicted rollout during the exploration, and higher $\lambda$ means otherwise. Similar to our experience on hardware, we find in our ablation that $\lambda = \sim 0.1$ produces the best behaviors, Fig. \ref{fig: ablations}.
\subsubsection{Prediction horizon}
Planning or prediction horizon plays a key role in MPC policies, especially in the absence of a good estimate of the cost-to-go. For standard MPC algorithms, longer prediction horizons are generally preferred as they often replace the need for an accurate cost-to-go estimate. Surprisingly, we find that the MPPI policy performs best with a $40-50$ timestep horizon (Fig \ref{fig: ablations}). We hypothesize that the fixed number of samples prevents our MPPI policy from sufficiently exploring the longer-horizon planning problem, leading to poor performance.
\subsubsection{Number samples}
Generally, more samples are preferred in sampling-based optimization algorithms, usually at the expense of more computation. To our surprise, the cost plateaus at $\sim 40$ samples for the quadruped walking task, even if more time is allowed to compute the policy in simulation. Our practical experience with the real robot also indicates that limited computing is much better spent on achieving a $\sim100$ Hz policy than additional sample evaluations during agile locomotion.



\begin{table}[h!]
\centering
\begin{tabular}{lccc}
\toprule
 & Offline & Whole-Body & Gradient-Free \\
 &Training&Contact&\\
\midrule
MPC& No & No & No \\
RL& Yes & Yes  & Yes \\
MPPI (ours) & No & Yes & Yes \\
\bottomrule
\end{tabular}
\caption{\label{tab:comp} Comparison of features across MPC~\cite{bishop2024reluqp, cleach2024fast, kuindersma2016optimization, bledt2018cheetah, rigo2023contact,de_vincenti2023centralized}, RL~\cite{lee2020learning, hwangbo2019learning,cheng2024parkour, pmlr-v205-margolis23a, pmlr-v164-rudin22a,hoeller2024anymal, ji2023driblebot}, and MPPI}
\vspace{-10mm}
\end{table}


\subsection{Comparisons to RL and MPC}
We qualitatively compare our MPPI policy against current MPC and RL algorithms for legged robots in Table \ref{tab:comp}. MPC policies generally do not require offline policy training as they leverage model gradients for real-time policy optimization. However, it is computationally intractable to include legged robot whole-body dynamics and collision models during real-time MPC policy evaluation. Simulation-based RL offers a gradient-free alternative that optimizes a neural network policy offline by collecting data using full-body dynamics and contact information in simulation. The MPPI policy (ours) is, so far, the only class of policy that can achieve policy optimization with whole-body dynamics and contact models without any offline training and can be fast enough to run directly in real time on legged robot hardware.
\section{Conclusions and Future Work}\label{sec: conclusions}
We present the first deployment of whole-body sampling-based MPC on real-world legged robots. By leveraging a performant and easily parallelizable modern robotics simulation engine, we can generate real-time contact-rich, whole-body motion plans that were previously only possible through manual pre-specification or offline policy training.
\subsection{Limitations}
Sampling-based MPC, and MPC in general, are fundamentally myopic, which means they can not see and therefore reason about interactions beyond the prediction horizon. Complementing sampling-based MPC with a global planner can enable solution of complicated long-horizon tasks. Additionally, these methods can only plan for what the simulator can simulate. Many important physical properties such as fluids ~\cite{lee2023aquarium}, soft materials~\cite{li2020incremental}, or complicated contact and friction interactions~\cite{elandt_2019} between objects cannot be accurately or efficiently simulated by common robotics physics engines. 
\subsection{Future Work}
Several important future research directions remain: First, we believe closing the sim-to-real and real-to-sim loop by estimating contact and friction parameters online can significantly improve real-world policy performance. Second, improving upon our baseline MPPI controller by sampling from more sophisticated distributions or over different model parameters can enable better performance in real-world settings.  Finally, we plan to integrate the robot hardware system with the MJPC interactive software~\cite{howell2022mjpc}. This can empower users to change task parameters and cost terms and observe changes to robot behavior in real time, enabling much more efficient and intuitive task design and cost tuning for model-based controllers.

\section*{ACKNOWLEDGMENTS}

The authors thank Taylor Howell for assistance with MuJoCo and Mitchell Fogelson, Swaminathan Gurumurthy, and Jon Arrizabalaga for valuable feedback and discussions on experiment design. 

\newpage
\bibliographystyle{IEEEtran}
\bibliography{references}

\end{document}